\documentclass[letterpaper]{article}
\usepackage{aaai15}
\usepackage{times}
\usepackage{helvet}
\usepackage{courier}
\usepackage{mydefs}
\usepackage{macros}

\begin{document}
\title{Generic Statistical Relational Entity Resolution in Knowledge Graphs\thanks{This manuscript was previously published as part of the doctoral dissertation, ``Probabilistic Models for Scalable Knowledge Graph Construction''}}
\author{Jay Pujara \\University of California, Santa Cruz\\Santa Cruz, CA 95064\\\email{jay@cs.umd.edu}
	\And Lise Getoor\\University of California, Santa Cruz\\Santa Cruz, CA 95064\\\email{getoor@soe.ucsc.edu}}

\maketitle
\begin{abstract}
Entity resolution, the problem of identifying the underlying entity of references found in data, has been researched for many decades in many communities. A common theme in this research has been the importance of incorporating relational features into the resolution process. Relational entity resolution is particularly important in knowledge graphs (KGs), which have a regular structure capturing entities and their interrelationships. We identify three major problems in KG entity resolution: (1) intra-KG reference ambiguity; (2) inter-KG reference ambiguity; and (3) ambiguity when extending KGs with new  facts. We implement a framework that generalizes across these three settings and exploits this regular structure of KGs. Our framework has many advantages over custom solutions widely deployed in industry, including collective inference, scalability, and interpretability. We apply our framework to two real-world KG entity resolution problems, ambiguity in NELL and merging data from Freebase and MusicBrainz, demonstrating the importance of relational features.
\end{abstract}
\section{Introduction}\label{sec:intro}

Entity resolution has been a longstanding challenge\cite{elmagarmid:tkde07} that has lead to significant research in many communities, including databases\cite{hernandez:sigmod95}, statistics\cite{winkler:tr06,fellegi:jass69}, information retrieval\cite{dong:sigmod05}, and natural language processing\cite{culotta:naacl07}. While entity resolution occurs in many settings, one setting that is particularly relevant to the current research landscape is entity resolution for knowledge graphs. In the past decade, a myriad of research projects in academia and industry have sought to automatically extract information from freely available text, images, video, and audio and assemble these extractions into entity-centric knowledge bases known as knowledge graphs.

In comparison to the general problem of entity resolution, knowledge graphs presents additional opportunities, complexities and challenges. We analyze two key facets of entity resolution problems arising from the structure of knowledge graphs: using knowledge graph features and supporting collective dependencies in co-reference judgments. We begin by identifying the problems confronting entity resolution in knowledge graphs and then develop a general model adaptable to many entity resolution tasks and scenarios.

The general problem of entity resolution is to take a set of references, such as proper names found in text or spoken language or bounding boxes found in images or video, and produce a mapping from these references to entities, which represent a single concept. This problem has two popular formulations: clustering or pairwise prediction.
When entity resolution is formulated as a clustering problem, the set of references are clustered, and each cluster of references represents an entity. In contrast, when entity resolution is formulated as a pairwise matching problem, each pair of references are assessed for equality and a connected component of equal references represents an entity. In both formulations, a key problem is measuring the similarity of references, either to determine cluster coherence or to produce pairwise co-reference predictions.

The earliest entity resolution research focused on developing specialized similarity measures for strings and attributes\cite{winkler:tr06}. More recent work in entity resolution has focused on using relationships between references to generate \textit{relational} features.  These relational features introduce dependencies between co-reference decisions for different references, resulting in a collective model that can outperform conventional approaches\cite{rastogi:vldb11,kalashnikov:sdm05,singla:icdm06}. For example, \cite{bhattacharya:tkdd07} introduce relational features and similarities, and using a collective relational clustering approach, demonstrate superior results to non-collective approaches. However, in many cases, these relational entity resolution models require cumbersome feature engineering and careful implementation that preserves scalability. One key requirement for knowledge graph entity resolution is the ability to translate knowledge graph features, such as attributes, types, and the many different relationships between entities, into features that can be used to determine the similarity of references.

A second key requirement for entity resolution in knowledge graphs is correctly handling collective dependencies in entity resolution decisions. Entity resolution problems are inherently collective due to transitivity or functionality constraints of equality. More formally, when resolving a set of references, a transitivity constraint requires that if \reference{A} and \reference{B} are co-referent, and \reference{B} and \reference{C} are co-referent, then \reference{A} and \reference{C} must also be co-referent. A functionality constraint can exist in a setting where a bijective mapping between references in two sets, \referenceset{S} and \referenceset{T}, is desired, if \reference{A} $\in$ \referenceset{S} and \reference{B} $\in$ \referenceset{T} are co-referent, then, for all \reference{C} $\in$ \referenceset{T}, \reference{A} and \reference{C} cannot be co-referent. While transitivity and functionality are standard examples of collective entity resolution challenges, the knowledge graph setting often includes more sophisticated examples of collective reasoning. For example, if we have two knowledge graphs that include references with relations pertaining to genealogical information, we might have references such as: \relpred(\eo, \oo, \parentrel), \relpred(\et, \ot, \parentrel), then determining that \eo\termspc and \et\termspc are co-referent can provide useful information that \oo\termspc and \ot\termspc are potentially co-referent as well.
	
\commentout{Another complication is that relational features may be dynamic, changing in response to new additions and inferences in the knowledge graph. \cite{namata:kdd11} considers the problem of recomputing relational features in response to entity resolution decisions, as well as the other graph identification tasks, node labeling and link prediction. In each case, relational features provide an important signal in making entity co-reference decisions.

A second difficulty is that entity resolution is inherently collective, as a result of transitivity or functionality properties of equality. For example, when resolving a set of references, if \reference{A} and \reference{B} are co-referent, and \reference{B} and \reference{C} are co-referent, then \reference{A} and \reference{C}. Or, in a setting where a bijective mapping between references in two sets \referenceset{S} and \referenceset{T} is desired, if \reference{A} $\in$ \referenceset{S} and \reference{B} $\in$ \referenceset{T} are co-referent, then, for all \reference{C} $\in$ \referenceset{T}, \reference{A} and \reference{C} cannot be co-referent.}

\section{Problem Definition}\label{sec:kgi_er:problem}
Our discussion hints at the diversity of entity resolution problems in knowledge graphs. Different phases of knowledge graph construction may face unique entity resolution challenges. We enumerate three general cases where entity resolution is necessary in knowledge graphs. Entity resolution may be required to:
\listspacing
\begin{enumerate}
	\item resolve ambiguity in a set of candidate extractions
	\item incorporate new extractions into an existing knowledge graph
	\item combine information from two or more knowledge graphs
\end{enumerate}
\end{spacing}
\noindent We discuss each of these scenarios in detail in the following paragraphs.

\subsection{Ambiguity In Candidate Extractions}\label{subsec:kgi_er:problem:ambiguous_extractions}
Knowledge graphs are commonly constructed by incorporating the outputs of information extraction methods. These information extraction techniques are subject to many sources of ambiguity. Each technique may process the same information differently, yielding many references from the same source material. Furthermore, the extraction source material may be inherently ambiguous, using different references for the same entity within a document, such as partial names or titles. Another common problem is anaphora, such as when a pronoun is used to with an ambiguous referent. Finally, the extractions are drawn from a corpus of documents, and each document may have variations in the representation of references, such as alternate spellings, prefixes, suffixes, and abbreviations. In addition to the noise in entity references, noise also exists in attributes and relations ascribed to each reference. In this scenario, the goal is to cluster a set of noisy references with noisy attributes and relations into a coherent set of entities. 

\subsection{Adding New Extractions to a Knowledge Graph}\label{subsec:kgi_er:problem:extend_graph}
A somewhat simpler problem is extending an existing knowledge graph using new extractions. In this setting, the goal is to map each reference to an existing entity in the knowledge graph, or introduce a new entity into the knowledge graph. One strategy for dealing with new entities that do not exist in the knowledge graph is skolemization, where each potential new entity is given a unique identifier. References can now be matched with existing entities or the new, skolemized entities in the knowledge graph, casting the problem into the well-studied task of surjective bipartite matching from references to entities. 

Through this formulation, the added constraint that each reference must match a single entity can often simplify the entity resolution process. While the attributes and relationships of the extracted reference may be noisy, as motivated in the previous scenario, the attributes and relationships of entities in the knowledge graph are expected to be highly reliable. As a result, relational features and attribute similarity play a more significant role in determining whether a reference can be resolved to an existing entity in the knowledge graph, or due to conflicting information, the reference should be added as a new entity with different attributes and relations.

\subsection{Combining Multiple Knowledge Graphs}\label{subsec:kgi_er:problem:multiple_graphs}
The final knowledge graph entity resolution scenario adheres most closely to the traditional approaches to entity resolution, where the goal is to combine information from two or more databases. In this setting, the goal is to find a mapping between entities in knowledge graphs, and then combine the attributes and relations of these entities. This problem can be formulated as mapping each knowledge graph to a ``canonical'' knowledge graph or instead be cast as a pairwise matching task between each pair of knowledge graphs. The latter formulation can introduce additional complexity in the form of transitivity constraints for equality across all knowledge graphs. These constraints can add new features for entity resolution, but may also make the desired mapping more computationally demanding. A further complication in this setting is that the knowledge graphs may use different schemas and ontologies. This problem is not covered in this work, but the development of standard ontologies and the problems of ontology matching or schema mapping have been extensively researched.

While these three entity resolution settings each present unique challenges, our goal is to provide a unified model for entity resolution. The goal of this model is to adapt to the diverse circumstances present in knowledge graph construction tasks. In the next section, we outline the structural elements of this model, and then introduce a probabilistic model for entity resolution that incorporates these elements into an entity resolution system.

\section{Approach}\label{sec:kgi_er:approach}
The crucial aspect that distinguishes knowledge graphs from standard entity resolution problems is the rich and regular structure of the knowledge graph, which provides relational features. Our goal is to leverage this structure to build an entity resolution model that is easy to understand and customize, while still capturing the rich information present in the knowledge graph. We consider two dimensions to the entity resolution model: feature granularity and collective inference. First, we organize the features in knowledge graphs based on the granularity of knowledge required. While the most basic features rely on string similarity or generic rules of functionality and transitivity, more complicated features involve new entities, attribute similarity, equivalence classes of relations, and domain-specific patterns. Each of these features can be classified as local (involving a single co-reference decision) or collective (imposing a dependency between two or more co-reference decisions). \tabref{tab:kgi_er:feature_summary} summarizes the knowledge graph features used by the entity resolution methods, and the following subsections delve more deeply into each of these feature sets. For each type of feature, we provide examples of corresponding logical rules. These rules can be combined in a probabilistic modeling framework, such as probabilistic soft logic, to produce a collective probabilistic graphical model for entity resolution.

\subsection{Local and Collective Knowledge Graph Features} \label{subsec:kgi_er:approach:kg_features}
As motivated earlier, there are two broad classes of features in knowledge graphs: local and collective. 
Local features are those that can be computed for a pair of entities (or references) independently of the co-reference decisions of other entities in the knowledge graph. Examples of local features include string similarity of names, image similarity of photographs, type agreement, and attribute agreement. One key characteristic for a local feature is that its value does not depend on the entity resolution decisions for other pairs of entities. This characteristic allows local features to be computed once for a pair of features and reused. Consequentially, relying on local features for entity resolution can decrease computational overhead and improve entity resolution performance.

In contrast to local features, collective features involve dependencies between co-reference decisions, and due to these dependencies are more difficult to compute. The transitivity and functionality constraints in the introduction are examples of common collective features that have been used in entity resolution. However, the structure of knowledge graphs allow many more collective features to be generated using relationships between entities. Knowledge graph features can be abstract, such as the overlap of object-arguments for a reference's relations, or very concrete, such as the link between parents and children in the earlier example. 

\begin{table*}[t]
	\centering
	\caption{Knowledge graph features categorized based on collective dependencies and level of granularity}
	\begin{tabular}{l cc cc}\toprule
						& \phantom{a} & local & \phantom{a} & collective \\ \midrule
		basic			& \phantom{a} & similarity scores & \phantom{a} & transitive, functional, sparsity \\
		new entity		& \phantom{a} & new entity prior & \phantom{a} & new entity penalty (sparsity) \\
		abstract KG		& \phantom{a} & type matching, type penalty & \phantom{a} & relation matching/equivalence \\
		domain-specific	& \phantom{a} & restricted type matching & \phantom{a} & restricted relation matching \\ \bottomrule
	\end{tabular}
	\label{tab:kgi_er:feature_summary}
\end{table*}

\subsection{Knowledge Graph Models at Different Granularity}\label{subsec:kgi_er:approach:kg_granularity}
In this section, we develop components for a knowledge graph entity resolution model. The components have been classified into four categories:
\listspacing
\begin{enumerate}
	\item {\bf basic} features that are common to all entity resolution scenarios
	\item {\bf new entity} features that helpful when adding new entities into a knowledge graph
	\item {\bf abstract KG} features that are universal across many knowledge graph structures
	\item {\bf domain-specific} features that are designed to resolve a particular class of entities
\end{enumerate}
\end{spacing}
In the subsequent sections, we will introduce logical rules for each type of feature, distinguishing between local and collective rules. The goal of these rules is to determine a pairwise resolution between two entities, denoted by \samepred(\eo, \et) for entities \eo and \et. Note that the \samepred\termspc predicate is distinct from the \sameentpred\termspc predicate, which is used to capture ontological information, such as \texttt{owl:sameAs}. 

Since knowledge graphs routinely contain millions of entities, assessing pairwise equality between all entities is infeasible. A common technique to avoid the polynomial explosion of entity matching is \textbf{blocking}, which uses a simple heuristic to produce potential entity matches. Using this smaller set of possible resolutions can substantially improve scalability. In the following rules, we will represent a blocked pair of entities with the predicate \candsamepred\termspc. Blocking can also be used to restrict matches based on the entity resolution scenario. For example, when incorporating new extractions into a knowledge graph, where the goal is to map references in a set of extractions to an existing knowledge graph, blocking can be used to scope entity resolution to only allow matches between extractions and the knowledge graph, disallowing matches within the extractions or within the knowledge graph.

\section{Modeling Knowledge Graph Entity Resolution}
\subsection{Basic Features}
\noindent\textbf{Rules for Local Features}\\
Basic features are those common to all entity resolution scenarios, such as similarity functions and prior probabilities. we introduce three rules corresponding to basic local features. \ruleref{rule:kgi_er:negprior} and \ruleref{rule:kgi_er:posprior} are priors for \samepred. Often, a negative prior (\ref{rule:kgi_er:negprior}) is useful to implement a default policy that entities are not co-referent unless supported by evidence. A positive prior can also be useful in some models to establish a baseline match confidence for two entities that have been blocked. 

The final basic local rule uses a similarity function, \simpred, to assess whether two entities are co-referent. In general, the similarity function can depend on the representation of the entities (e.g. images, sound files, or textual representations). A great deal of research in entity resolution has been devoted to designing effective similarity functions for entity resolution. Examples of popular similarity functions are Levenstein \citep{navarro:acm01,wagner:jacm74}, Jaro \citep{jaro:sim95}, Jaro-Winkler \citep{winkler:census99}, Monge-Elkan \citep{elkan:kdd96}, Fellegi-Sunter \citep{fellegi:jasa69}, Needleman-Wunsch, and Smith-Waterman \citep{durbin:book98}. In \ruleref{rule:kgi_er:similarity} the similarity function is not explicitly specified, but a popular similarity function or combination of functions \citep{bilenko:kdd03} can be used to determine similarity.
\rulespacing\begin{align}
&&&\pslneg \samepred(\aeo, \aet)\label{rule:kgi_er:negprior}\\
\alignbreak
&&\candsamepred(\aeo,\aet)\termspc\termspc&\notag\\
&&& \pslthenspc \samepred(\aeo, \aet)\label{rule:kgi_er:posprior}\\
\alignbreak
&&\candsamepred(\aeo,\aet)\termspc\termspc& \pslandspc  \simpred(\aeo,\aet)\notag\\ 
&&&\pslthen \samepred(\aeo, \aet)\label{rule:kgi_er:similarity}
\end{align}\end{spacing}

\noindent\textbf{Rules for Collective Features}\\
The collective basic features incorporate the fundamental properties of equality: symmetry (\ruleref{rule:kgi_er:symmetry}) and transitivity (\ruleref{rule:kgi_er:transitivity}). Symmetry enforces the constraint that the order of the arguments to \samepred\termspc do not matter. Transitivity, discussed in the introduction, ensures that the co-reference process generates tight clusters of entities by encouraging co-reference cliques. Finally, \ruleref{rule:kgi_er:sparsity} has an opposite effect, encouraging sparsity by promoting functionality for the \samepred\termspc predicate. Not all entity resolution scenarios require functionality for co-references, but those discussed in the sections on extending a knowledge and combining knowledge graphs can benefit from such constraints.

\rulespacing\begin{align}
&&\samepred(\aeo, \aet)\termspc\termspc &&\notag\\
&&& \pslthenspc \samepred(\aet, \aeo)&\label{rule:kgi_er:symmetry}\\
\alignbreak
&& \candsamepred(A,B)\termspc\termspc &\pslandspc \candsamepred(B,C)&\notag\\ 
&& \pslandspc \candsamepred(A,C)\termspc\termspc&\pslandspc 
\samepred(A,B)&\notag\\ 
&&& \pslandspc \samepred(B,C)&\notag\\ 
&&&\pslthenspc \samepred(A,C)&\label{rule:kgi_er:transitivity}\\
\alignbreak
&&\candsamepred(A,B)\termspc\termspc &\pslandspc \candsamepred(A,C)&\notag\\ 
&&& \pslandspc \samepred(A,B)& \notag\\
&&& \pslthenspc  \pslneg\samepred(A,C)&\label{rule:kgi_er:sparsity}
\end{align}\end{spacing}%
\subsection{New Entity Features} 
\noindent\textbf{Rules for Local Features}\\
In problem settings where entity resolution is matching with respect to an existing knowledge graph, such as extending a knowledge graph and merging multiple knowledge graphs, the appropriate entity may not exist in the target knowledge graph. In these settings, we generate a new entity placeholder for each source reference. This placeholder will have no inherent relations, types, or attributes and will have a default similarity value. we designate these entities using the \newentpred\termspc predicate. \ruleref{rule:kgi_er:new_entity_prior} establishes a prior that any reference matches a new entity. In subsequent rules, the \newentpred\termspc will be used to  scope the rule to existing entities, which avoids penalizing new entity matches based on relations, types and attributes which are missing.
\rulespacing\begin{align}
&&\candsamepred(\aeo,\aet)\termspc\termspc &\pslandspc \newentpred(\aeo)&\notag\\
&&& \pslthenspc \samepred(\aeo,\aet)& \label{rule:kgi_er:new_entity_prior}
\end{align}\end{spacing}

\noindent\textbf{Rules for Collective Features}\\
While a prior can be helpful, the desired behavior in entity resolution systems is to add a new entity only when no other entity in the knowledge graph appears to match. \ruleref{rule:kgi_er:new_entity_sparsity} prevents a new entity from matching when a previously existing entity is a strong match for a reference.
\rulespacing\begin{align}
&&\samepred(\aeo,\aet)\termspc\termspc &\pslandspc \candsamepred(\aeo,\aeth)& \notag\\
&& &\pslandspc \newentpred(\aeth)& \notag\\
&&&  \pslthenspc \pslneg\samepred(\aeo,\aeth)&\label{rule:kgi_er:new_entity_sparsity}
\end{align}\end{spacing}%
\subsection{Abstract Knowledge Graph Features}
Abstract knowledge graph features use the relational structure and attributes shared by all knowledge graphs. The key strength is that these features are broadly applicable to any knowledge graph entity resolution problem. In scenarios such as disambiguating references within a knowledge graph, abstract knowledge graph rules can be used to collectively infer relations and \labels in the knowledge graph while simultaneously determining entity co-references. However, one drawback of abstract knowledge graph rules is that their broad applicability may limit their usefulness. Rules that are agnostic to the particular \labels and relations in a knowledge graph may have difficulty prioritizing which \labels and relations are useful for entity resolution. One potential solution to this issue when ample training data is available is to introduce rules and then learn rule weights for each \labell and relation separately. 

\noindent\textbf{Rules for Local Features}\\
Knowledge graph entities have associated properties such as attributes, labels, and type information that provide the basis for local features. \ruleref{rule:kgi_er:label_agreement} specifies that these properties agree for two entities. Since many potential candidate matches may share the same properties, the rule is mediated by the candidate similarity, supporting similar matches more strongly than dissimilar matches. While entities with agreeing properties are a signal of co-reference, properties that are missing or explicitly disagree can be strong signals against co-reference. \ruleref{rule:kgi_er:label_disagreement} requires that co-referent entities share properties, but provides an exception for new entities which lack properties. Note that a symmetric rule for the second entity is not shown. These rules are most useful in entity resolution settings where knowledge graph information is relatively complete, whereas noisy or incomplete extractions may hamper entity resolution. 
\ruleref{rule:kgi_er:label_mutex} provides a stronger signal by incorporating the knowledge graph ontology, disallowing entities with mutually-exclusive properties from matching. Even when extractions are noisy and properties incomplete, this signal can provide strong evidence against a potential co-reference match.
\rulespacing\begin{align}
	&&\candsamepred(\aeo,\aet)\termspc\termspc&\pslandspc\simpred(\aeo,\aet)&\notag\\  
	&&\pslandspc\catpred(\aeo,\cat)\termspc\termspc&\pslandspc\catpred(\aet,\cat)&\notag\\
	&&&\pslthenspc\samepred(\aeo,\aet)&\label{rule:kgi_er:label_agreement}\\
	\alignbreak
	&&\candsamepred(\aeo,\aet)\termspc\termspc  &\pslandspc\catpred(\aeo,\cat)&\notag\\
	&&\pslandspc\pslneg\catpred(\aet,\cat)\termspc\termspc&\pslandspc\pslneg\newentpred(\aet)&\notag\\ 
	&&&\pslthenspc\pslneg\samepred(\aeo,\aet)&\label{rule:kgi_er:label_disagreement}\\
	\alignbreak
	&&\candsamepred(\aeo,\aet)\termspc\termspc  &\pslandspc\catpred(\aeo,\acato)&\notag\\ 
	&& \pslandspc\catpred(\aet,\acatt)\termspc\termspc& \pslandspc\mutex(\acato,\acatt)&\notag\\
	&&& \pslthenspc\pslneg\samepred(\aeo,\aet)&\label{rule:kgi_er:label_mutex}
\end{align}\end{spacing}

\noindent\textbf{Rules for Collective Features}\\
The collective abstract knowledge graph entity resolution parallel the local rules, but operate over relations and involve pairs of co-referent entities. \ruleref{rule:kgi_er:relational_agreement} requires that two co-referent entities have the same relation with co-referent objects. The collective nature of the rule introduces a dependence between entities that participate in the same relation across knowledge graphs, supporting co-references between the subjects and objects of the relation. \ruleref{rule:kgi_er:relational_disagreement} has the opposite effect, penalizing co-references for matches between existing entities that do not share the same relations. Echoing the previous remarks on knowledge graph rules, these rules have limited usefulness in noisy or incomplete knowledge graphs where many relations may be missing. However, \ruleref{rule:kgi_er:relational_mutex} uses the ontology to find a stronger signal in mutually-exclusive relations.
\rulespacing\begin{align}
	&&\candsamepred(\aeo,\aet)\termspc\termspc  &\pslandspc\candsamepred(\aoo,\aot)& \notag \\
	&&\pslandspc\simpred(\aeo,\aet)\termspc\termspc &\pslandspc\samepred(\aoo,\aot)& \notag\\  
	&&\pslandspc\relpred(\aeo,\aoo, \rel)\termspc\termspc&\pslandspc\relpred(\aet,\aot, \rel)& \notag\\  
	&&&\pslthenspc\samepred(\aeo,\aet)&\label{rule:kgi_er:relational_agreement}\\
	\alignbreak
	&&\candsamepred(\aeo,\aet)\termspc\termspc  &\pslandspc\candsamepred(\aoo,\aot)& \notag\\
	&&\pslandspc\samepred(\aoo,\aot)\termspc\termspc
	&\pslandspc\pslneg\relpred(\aeo,\aoo, \rel)& \notag\\  
	&&\pslandspc\pslneg\newentpred(\aeo)\termspc\termspc &\pslandspc\pslneg\newentpred(\aoo)&\notag\\		
	&&&\pslandspc\relpred(\aet,\aot, \rel)& \notag\\  
	&&&\pslthenspc\pslneg\samepred(\aeo,\aet)&\label{rule:kgi_er:relational_disagreement}\\
	\alignbreak
	&&\candsamepred(\aeo,\aet)\termspc\termspc  &\pslandspc\candsamepred(\aoo,\aot)& \notag\\
	&&\pslandspc\samepred(\aoo,\aot)\termspc&\pslandspc\relpred(\aeo,\aoo, \rel)& \notag\\  
	&&\pslandspc\relpred(\aet,\aot, \orel)\termspc &\pslandspc\rmutex(\rel,\orel)& \notag\\
	&&&\pslthenspc\pslneg\samepred(\aeo,\aet)&\label{rule:kgi_er:relational_mutex}
\end{align}\end{spacing}
	
\subsection{Domain-Specific Knowledge Graph Features}
While abstract knowledge graph features provide a generally-applicable tool for knowledge graph entity resolution, in many cases domain experts can rely on experience to assess the most important features for matching knowledge graphs. Since our model uses interpretable rules that are easy to generate, domain experts can easily add and remove rules to the model to capture the most relevant relationships. In this section, we provide some example rules for the task of matching knowledge graphs in the music domain. These rules are derived from rules used in an industry knowledge graph matching system, supporting the assertion that rules are a natural and common form of supplying domain expertise for knowledge graphs.

\noindent\textbf{Rules for Local Features}\\
One example of a domain rule that strongly supports co-reference are relations with categorical domains. The \releasetype\termspc relation in musical domains differentiates between singles, EPs, and albums. Since the domain of the relation is a small, enumerated set, matching release types across co-references is important. \ruleref{rule:kgi_er:domain_releasetype} incorporates this domain knowledge in a rule for release type matching. Just as some relations are more important than others, so are types, attributes and labels. While general purpose ontologies have a \person\termspc type, a more specific type can be far more useful in matching. \ruleref{rule:kgi_er:domain_artist} provides a special case for \artist, a subtype of \person. One way of interpreting this rule is a type-specific prior for entity matches. By choosing appropriate weights, these rules can also moderate the importance of a similarity metric in a particular domain. For example, a high similarity value may not be meaningful for a broad domain (e.g. \person) but can provide a stronger disambiguating signal for a more selective domain (e.g. \artist).

\rulespacing\begin{align}
	&&\candsamepred(\aeo,\aet)\termspc\termspc &\pslandspc \simpred(\aeo, \aet)&\notag\\
	&&&\pslandspc \relpred(\aeo, \cat, \releasetype)&    	\notag\\  
	&&&\pslandspc \relpred(\aet, \cat, \releasetype)& \notag\\  
	&&&\pslthenspc \samepred(\aeo,\aet)&\label{rule:kgi_er:domain_releasetype}\\
	\alignbreak
	&&\candsamepred(\aeo,\aet)\termspc\termspc  &\pslandspc \simpred(\aeo, \aet)&\notag\\
	&&&\pslandspc \catpred(\aeo, \artist)& \notag\\
	&&&\pslandspc \catpred(\aet, \artist)&\notag\\    
	&&&\pslthenspc \samepred(\aeo,\aet)&\label{rule:kgi_er:domain_artist}
\end{align}\end{spacing}

\noindent\textbf{Rules for Collective Features}\\
Similarly, domain experts can select the most important relations for resolution in a domain. \ruleref{rule:kgi_er:release_album} which focuses on co-referent releases of the same album can be more useful than a rule which focuses on \releaselabel\termspc since a label typically has many releases. Domain rules can also incorporate more complex criteria. \ruleref{rule:kgi_er:domain_multilevel} has a similar form to normal collective relational rules, but includes an additional constraint that the albums and artists must all come from the same genre.
\rulespacing\begin{align}
	&&\candsamepred(\aeo,\aet)\termspc\termspc  &\pslandspc \simpred(\aeo, \aet)&\notag\\
	&&\pslandspc \candsamepred(\aoo,\aot)\termspc\termspc &\pslandspc	\samepred(\aet,\aeo)& \notag\\  
	&&&\pslandspc	\relpred(\aeo, \aoo, \releasealbum)&  	\notag\\  
	&&&\pslandspc	\relpred(\aet, \aot, \releasealbum)&  	\notag\\  
	&&&\pslthenspc\samepred(\aoo,\aot)&\label{rule:kgi_er:release_album}\\
\alignbreak
	&&\candsamepred(\aeo,\aet)\termspc\termspc  & \pslandspc \candsamepred(\aoo,\aot)&	\notag\\
	&&\pslandspc \simpred(\aeo, \aet)\termspc\termspc &\pslandspc \simpred(\aoo, \aot)&	\notag\\
	&&&\pslandspc	\relpred(\aeo, \aoo, \albumartist)&   	\notag\\  
	&&&\pslandspc	\relpred(\aet, \aot, \albumartist)& 	\notag\\  
	&&&\pslandspc	\relpred(\aeo, G, \albumgenre)& 	\notag\\  
	&&&\pslandspc	\relpred(\aet, G, \albumgenre)& 	\notag\\  
	&&&\pslandspc	\relpred(\aoo, G, \artistgenre)& 	\notag\\  
	&&&\pslandspc	\relpred(\aot, G, \artistgenre)& 	\notag\\  
	&&&\pslandspc	\samepred(\aoo, \aot)& 	\notag\\  
	&&&\pslthenspc\samepred(\aeo,\aet)&\label{rule:kgi_er:domain_multilevel}
\end{align}\end{spacing}%

\begin{table*}[ht!]
	\centering
	\caption{Summarizing entity resolution rules and matching them to application}
	\begin{tabularx}{\textwidth}{X cc cc cc cc}\toprule
		& \phantom{ab} & Local/ & \phantom{ab} & New & \phantom{ab} & Extend & \phantom{ab} & Multiple\\ 
		& \phantom{ab} & collective & \phantom{ab} & extractions & \phantom{ab} & KG & \phantom{ab} & KGs\\ 
		\midrule
	Negative prior (\ref{rule:kgi_er:negprior}) 
	& \phantom{ab} & L & \phantom{ab} & Y & \phantom{ab} & Y & \phantom{ab} & Y\\
	Positive prior (\ref{rule:kgi_er:posprior})
	& \phantom{ab} & L & \phantom{ab} & Y & \phantom{ab} & Y & \phantom{ab} & Y\\
	Similarities (\ref{rule:kgi_er:similarity})
	& \phantom{ab} & L & \phantom{ab} & Y & \phantom{ab} & Y & \phantom{ab} & Y\\
	Symmetry (\ref{rule:kgi_er:symmetry})
	& \phantom{ab} & C & \phantom{ab} & Y & \phantom{ab} & Y & \phantom{ab} & Y\\
	Transitivity (\ref{rule:kgi_er:transitivity})
	& \phantom{ab} & C & \phantom{ab} & Y & \phantom{ab} & N & \phantom{ab} & Y\\
	Sparsity (\ref{rule:kgi_er:sparsity})
	& \phantom{ab} & C & \phantom{ab} & N & \phantom{ab} & N & \phantom{ab} & Y\\
	New Entity prior (\ref{rule:kgi_er:new_entity_prior})
	& \phantom{ab} & L & \phantom{ab} & N & \phantom{ab} & Y & \phantom{ab} & Y\\
	New Entity penalty (\ref{rule:kgi_er:new_entity_sparsity})
	& \phantom{ab} & C & \phantom{ab} & N & \phantom{ab} & Y & \phantom{ab} & Y\\
	Label agreement (\ref{rule:kgi_er:label_agreement})
	& \phantom{ab} & L & \phantom{ab} & N? & \phantom{ab} & Y & \phantom{ab} & Y\\
	Label disagreement (\ref{rule:kgi_er:label_disagreement})
	& \phantom{ab} & L & \phantom{ab} & N? & \phantom{ab} & Y? & \phantom{ab} & Y?\\
	Label exclusion (\ref{rule:kgi_er:label_mutex})
	& \phantom{ab} & L & \phantom{ab} & Y & \phantom{ab} & Y & \phantom{ab} & Y\\
	Relational agreement	(\ref{rule:kgi_er:relational_agreement})
	& \phantom{ab} & C & \phantom{ab} & N? & \phantom{ab} & Y & \phantom{ab} & Y\\
	Relational disagreement (\ref{rule:kgi_er:relational_disagreement})
	& \phantom{ab} & C & \phantom{ab} & N? & \phantom{ab} & Y? & \phantom{ab} & Y?\\
	Relational exclusion (\ref{rule:kgi_er:relational_mutex})
	& \phantom{ab} & C & \phantom{ab} & Y & \phantom{ab} & Y & \phantom{ab} & Y\\
	Domain-specific categorical relations (\ref{rule:kgi_er:domain_releasetype})
	& \phantom{ab} & L & \phantom{ab} & Y & \phantom{ab} & Y & \phantom{ab} & Y\\
	Domain-specific prior (\ref{rule:kgi_er:domain_artist})
	& \phantom{ab} & L & \phantom{ab} & Y & \phantom{ab} & Y & \phantom{ab} & Y\\
	Domain-specific relations (\ref{rule:kgi_er:release_album})
	& \phantom{ab} & C & \phantom{ab} & Y? & \phantom{ab} & Y & \phantom{ab} & Y\\
	Domain-specific relational criteria 	(\ref{rule:kgi_er:domain_multilevel})
	& \phantom{ab} & C & \phantom{ab} & Y? & \phantom{ab} & Y & \phantom{ab} & Y\\
	\bottomrule
	\end{tabularx}
	\label{tab:kgi_er:rule_summary}
\end{table*}
\subsection{Synthesis}
The previous section introduced a number of rules for entity resolution, categorized by whether the rule used local or collective information and the granularity of the knowledge graph features used. In the discussion of each rule, we referenced the three knowledge graph entity resolution scenarios and the conditions under which the rule was applicable to the scenario. The rules and this discussion is summarized in \tabref{tab:kgi_er:rule_summary}. Note that some of the entries have question marks, which reinforce the guidance that the corresponding rules may be appropriate based on dataset characteristics such as noise and sparsity. 

The knowledge graph entity resolution model presented in this section is a general and versatile approach to entity resolution in richly structured domains. Since the requirements of different entity resolution scenarios vary, care must be taken to select the appropriate rules and design meaningful domain-specific rules. However the proliferation of domain-specific entity resolution methods \citep{durbin:book98,winkler:tr06} and anecdotal evidence from many projects in industry suggest that many bespoke entity resolution systems are already in use. Despite the widespread use of such systems and substantial research in entity resolution, no general-purpose, collective framework has been adopted across domains.

This work provides a general guide to designing entity resolution systems for knowledge graphs. The rules presented can be used as templates for many approaches that jointly model entity resolution decisions, such as linear programs and probabilistic models. 
We use the rules as the basis for a probabilistic soft logic (PSL) program for performing entity resolution. PSL is a natural choice for entity resolution models, since entity resolution models have many collective dependencies, use real-valued similarity measures, and often include a vast number of entities. 
\commentout{
\section{Approach}
\begin{itemize}
	\item Formal problem definition
\item Define a model for entity resolution using series of PSL rules
\item 3.1: General non-collective entity resolution rules (Basic)
\begin{itemize}
\item Confidence from source(s)
\item New Entity priors
\item KG-relevant non-collective entity resolution rules
\item type-specific prior
\end{itemize}
\item 3.2 General collective entity resolution rules
\begin{itemize}
\item Sparsity in resolution
\item Preference for existing entities over new entities
\item Relational similarity (Rel(R,A,B) \& Rel(R,A,C) \& Same(A,B))
\end{itemize}
\item 3.3 Domain-specific, non-collective entity resolution rules
\begin{itemize}
\item type-based priors
\end{itemize}
\item 3.4 Domain-specific, collective entity resolution rules
\begin{itemize}
\item Ontological relationships from schema
\end{itemize}
\item With training data, learn weights for rules
\item Use rules to define probability distribution over possible resolution pairs
\item Efficient inference provides resolution decision for full KG
\end{itemize}
}

\section{Evaluation}\label{sec:kgi_er:evaluation}
We evaluate our knowledge graph entity resolution approach on two very different datasets from different entity resolution scenarios. The first dataset, corresponding to the scenario where references in ambiguous extractions are resolved, involves clustering unresolved references with associated relations and attributes from different web sources. The second dataset, corresponding to the scenario in merging multiple knowledge graphs, involves resolving entities between the MusicBrainz music knowledge graph and the Freebase broad-coverage knowledge graph.
\subsubsection*{NELL}
NELL extracts a series of facts from text, and uses a set of heuristics to map textual references to entities. This entity mapping process includes two phases: first, textual references are clustered to identify senses and then these textual references are mapped to the appropriate senses. The entity mapping process does not use the context of the knowledge graph, which can improve the performance on entity mapping. Furthermore, the entity mapping process does not attempt to perform entity resolution between different textual references that refer to the same underlying entity.

In order to investigate the effectiveness of entity resolution applied to ambiguous candidate extractions, We worked with the NELL team to collect data from a new NELL instance that performed only the first phase of entity mapping, clustering textual references to generate senses. The second phase of entity mapping was not performed, so this NELL instance produced raw candidate extractions in terms of the original textual references. The goal in this setting is to collectively determine the facts in the knowledge graph along with the entity co-references. 

NELL's Entity Resolver produces match scores for pairs of textual references. We extend these match scores by computing a number of string similarity metrics for each pair of textual references, using the SecondString library \citep{cohen:ijcaiws03}. The set of string similarities includes the Jaccard overlap (of characters), Jaro, Jaro-Winkler, Levenshtein, Monge-Elkan, and Smith-Wasserman similarity functions. These string similarities constituted local features for entity resolution.

Using data from the first iteration of NELL yielded 145K candidate relations, 200K candidate labels, 170K unique textual references that mapped to 190K potential entities. The NELL EntityResolver candidate generation produced 4K potential entity co-references.
Since the dataset was collected from a new NELL instance, no existing entity match information was used or available. Furthermore, since there were no pre-existing entities, each textual reference was considered unknown and no special handling of new entities was required.

NELL does not have a reliable source of entity resolution data, so we manually labeled entity co-references. For each method, we chose the top 50 as well as 50 randomly selected entity co-references from each method. This selection process yielded 421 co-references after duplicates were removed. We then removed the truth values and randomized the order of the co-references for judging. 

Entities were judged to be co-referent when there was an unambiguous interpretation of the textual references that corresponded to one entity. This, for example, excludes ``Giants'' matching ``San Jose Giants'' since many other sports teams share the same name. Similarly, when a textual reference was the amalgamation of two entities, matches with either entity were disallowed. For example, this invalidates ``Quito'' from matching with ``Quito and Cuenca''. However, merged entities were judged co-referent, allowing ``Konica'' and ``Konica Minolta'' to be co-referent since the company Konica merged with Minolta to become the merged company.

Results for the NELL entity resolution are reported in \tabref{tab:kgi_er:nell_results_overall}. The baseline, Basic-Local entity resolution uses only priors and the various similarity metrics.  

\begin{table*}[t]
	\begin{tabularx}{\textwidth}{Xc cc cc cc cc}\toprule
		Method & \phantom{ab} & AUPRC & \phantom{ab} & F1 & \phantom{ab} & Prec. & \phantom{ab} & Recall & \phantom{ab}\\\midrule
		Basic, Local & \phantom{ab} & 0.267 & \phantom{ab} & 0.333 & \phantom{ab} & 0.214 & \phantom{ab} & 0.759 & \phantom{ab}\\
		Basic \& KG, Local & \phantom{ab} & 0.247 & \phantom{ab} & 0.426 & \phantom{ab} & 0.298 & \phantom{ab} & 0.747 & \phantom{ab}\\
		Basic, All & \phantom{ab} & 0.307 & \phantom{ab} & 0.446 & \phantom{ab} & 0.333 & \phantom{ab} & 0.675 & \phantom{ab}\\
		Basic \& KG, All & \phantom{ab} & {\bf 0.351} & \phantom{ab} & {\bf 0.453} & \phantom{ab} & 0.383 & \phantom{ab} & 0.554 & \phantom{ab}\\\bottomrule
	\end{tabularx}			
	\caption{Comparing the performance of knowledge graph entity resolution rules in for the NELL dataset. Performance improves by adding knowledge graph features and collective features, with the best performance with both.}	
	\label{tab:kgi_er:nell_results_overall}
	%
\end{table*}

\subsubsection*{MusicBrainz and Freebase}
The second dataset for entity resolution involved mapping entities between two knowledge graphs. The first knowledge graph was from the MusicBrainz music knowledge base, available courtesy of the LinkedBrainz project.\footnote{\url{http://linkedbrainz.org/}}
The second knowledge graph was the publicly available Freebase knowledge base. 

An existing, proprietary pipeline to map entities between these two knowledge graphs exists. The pipeline uses Boolean rules restricted to discrete features. The system is designed to consider entity resolutions sequentially, and as a result cannot use all collective information between resolution decisions. When a match decision for an entity cannot be made by the pipeline, the entity is manually resolved by a human annotator. Evaluation of the existing pipeline showed a high error rate, while manually annotated entities contained no errors. Our experiments focus on the entities that were not successfully matched using the existing pipeline, which constitute the most difficult entity resolution decisions.

We begin with a dataset of 11K entities added to the MusicBrainz knowledge graph between 5/5/2014 and 6/29/14 that were manually annotated and have reliable ground truth. 
We identify 332K Freebase entities that are potential candidate matches for the MusicBrainz entities using a string similarity measure that is normalized for entity frequency. Since these newly added entities are often not found in Freebase, we generate new entity candidates for each MusicBrainz entity. The entity resolution dataset also includes 1M known entity mappings between the two knowledge graphs and 15.7M relations between entities. 

\begin{table*}[t]
	\begin{tabularx}{\textwidth}{Xc cc cc cc cc}\toprule
		Method & \phantom{ab}  & AUPRC & \phantom{ab} & F1 & \phantom{ab} & F1 (Exist) & \phantom{ab} & F1 (New) & \phantom{ab} \\\midrule
		Basic \& NewEntity, Local & \phantom{ab} &  0.416 & \phantom{ab} & 0.734 & \phantom{ab} & 0.169 & \phantom{ab} & 0.744 & \phantom{ab}\\
		Basic \& Domain, All; NewEntity, Local & \phantom{ab} & 0.569 & \phantom{ab} & 0.805 & \phantom{ab} & {\bf 0.305} & \phantom{ab} & 0.831 & \phantom{ab}\\
		Basic \& Domain \& NewEntity, All & \phantom{ab} & {\bf 0.724} & \phantom{ab} & {\bf 0.840} & \phantom{ab} & 0.070 & \phantom{ab} & {\bf 0.895} & \phantom{ab}\\\bottomrule
	\end{tabularx}
	\caption{Comparing the performance of knowledge graph entity resolution rules when merging MusicBrainz entities into the Knowledge Graph. Due to a skew toward new entities, the collective new entity rules dramatically improve overall performance, but with a substantial drop in the F1 measure for existing entities}
	\label{tab:kgi_er:mbz_kg_results}
\end{table*}

Table \ref{tab:kgi_er:mbz_kg_results} summarizes the results of these experiments. The baseline method uses only local rules, and achieves an area under the precision-recall curve (AUPRC) of 0.416 and an F1 measure of 0.734. Adding collective rules and domain-specific features that use the knowledge graph improves performance, with an AUPRC of 0.569 and an F1 of 0.805. Incorporating rules to handle new entities further improves performance with an AUPRC of 0.724 and an F1 measure of 0.840. 

To better understand the performance, we separate the F1 measure for existing entities and new entities. In the dataset we collected, the entity mappings are skewed toward new entities, so that approximately 75\% of entities in the MusicBrainz knowledge graph are not found in the Freebase entities. Thus the New Entity rules can have a dramatic influence on the performance by improving the performance on new entities while having a marked decrease in performance in existing entities. 

\section{Discussion}
The growing importance of knowledge graphs has resulted in an increased emphasis on entity resolution for such structured domains. The collective relationships in a knowledge graph provide the key to improving the performance of entity resolution. In this paper, we provided an inventory of important features necessary for entity resolution in the context of knowledge graphs and identified the corresponding knowledge graph settings where these features are important. Building entity resolution models, particularly collective models, has required a great deal of time and effort. The general nature of this model makes it applicable to many different problem settings, and a PSL implementation of our entity resolution model makes it accessible for rapid prototyping and experimentation for a variety of entity resolution problems.

\section{Acknowledgements}
This manuscript is an adaptation of a chapter of the dissertation ``Probabilistic Models for Scalable Knowledge Graph Construction'' \cite{pujara:thesis16}. Data for disambiguating NELL extractions was collected with the assistance of Bryan Kisiel, Jayant Krishnamurthy, and William Cohen. Work on entity resolution for Freebase data was undertaken during an internship with the Google Knowledge Vault team, and with the help of Luna Dong, Curtis Janssen, and Kevin Murphy.
\bibliography{pujara-starai16}
\bibliographystyle{aaai}
\end{document}